\documentclass[a4paper, 10pt, conference]{ieeeconf} 

\IEEEoverridecommandlockouts                              

\usepackage[scaled]{helvet} 
\usepackage{customstyle}

\title{\LARGE \bf
Meta-control of Dialogue Systems Using Large Language Models
}

\author{Kotaro Shukuri$^{1}$, Ryoma Ishigaki$^{1}$, Jundai Suzuki$^{1}$, Tsubasa Naganuma$^{1}$, Takuma Fujimoto$^{1}$ \\ Daisuke Kawakubo$^{1}$, Masaki Shuzo$^{1}$ and Eisaku Maeda$^{1}$
\thanks{*This work was supported by Grant-in-Aid for Scientific Research on Innovative Areas, Grant Numbers JP19H05693}
\thanks{$^{1}$K. Shukuri, R. Ishigaki, J. Suzuki, T. Naganuma, T. Fujimoto, D. Kawakubo, M. Shuzo, E. Maeda are with Tokyo Denki University,
        5 Senju Asahi-cho, Adachi-ku, Tokyo 120-8551, Japan
        {\tt\small \{20aj076@ms, 20aj012@ms, 20aj078@ms, 20aj098@ms, 20aj112@ms, 22amj10@ms, shuzo@mail, maeda.e@mail\}.dendai.ac.jp}}%
}
 
\begin{document}
\maketitle
\thispagestyle{empty}
\pagestyle{empty}

\begin{abstract}
Utilizing Large Language Models (LLMs) facilitates the creation of flexible and natural dialogues, a task that has been challenging with traditional rule-based dialogue systems. 
However, LLMs also have the potential to produce unexpected responses, which may not align with the intentions of dialogue system designers. To address this issue, this paper introduces a meta-control method that employs LLMs to develop more stable and adaptable dialogue systems. 
The method includes dialogue flow control to ensure that utterances conform to predefined scenarios and turn-taking control to foster natural dialogues. 
Furthermore, we have implemented a dialogue system that utilizes this meta-control strategy and verified that the dialogue system utilizing meta-control operates as intended.
%
\end{abstract}


\section{INTRODUCTION}
In the construction of dialogue systems, 
it is common to design rule-based systems to achieve system-driven dialogue tailored to their intended purpose. 
However, it was difficult for this rule-based dialogue system to make a dialogue like a chatting dialogue continue for a long period of time without breaking down. 
Recently, there has been an increasing trend towards utilizing Large Language Models (LLMs) to enable flexible and diverse utterance generation. 
However, controlling LLMs outputs explicitly remain a challenge, and creating stable and ongoing dialogue systems using LLMs is not always straightforward. 

The Dialogue Robot Competition (DRC2023) \cite{DRC2023} presented a challenge: constructing a multimodal dialogue system for an android robot acting as a travel agency representative. 
The objective of this dialogue is to attentively listen to user's travel needs, and formulate and present a satisfying travel plan. 
In addition to maintaining a stable dialogue that progresses toward a clear, predefined goal, it is necessary to have the ability to generate flexible utterances and control the dialogue in response to the user's various utterances. 



\section{OVERVIEW OF DIALOGUE SYSTEM DSML-TDU}

\begin{figure}[t]
 \centering
  \includegraphics[width= 80mm]{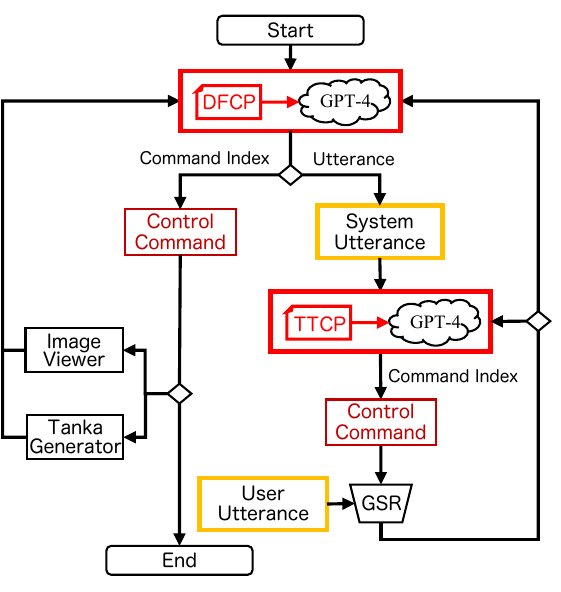}
 \caption{Overview of dialogue flow in DSML-TDU system}
 \label{f:flow}
\end{figure}

For DRC2023, we developed the DSML-TDU dialogue system, 
the overview of which is shown in Fig. \ref{f:flow}.
In managing the introduction, closing, and the 10-minute dialogue duration, 
we employed a rule-based dialogue control.
For other components of the system, 
we utilized an LLM for both utterance generation and dialogue control. 
We used GPT-4 as the LLM and Google Speech Recognizer (GSR) for speech recognition.
To effectively harness GPT-4, 
we prepared two types of prompts: 
``Dialogue Flow Control Prompt (DFCP)'' and 
``Turn-Take Control Prompt (TTCP).''
The distinctive features of these prompts are that, 
in addition to generating utterances, 
they facilitate branching decisions at scenario junctions 
and control the turn-taking in dialogues. 
We refer to this functionality as `meta-control.'
The prompts in this paper are originally written in Japanese and have been translated into English.


\section{META-CONTROL BY 2 PROMPTS FOR GPT-4}
\subsection{Dialogue Flow Control Prompt (DFCP)}

The DFCP is a tool designed to simultaneously generate utterances and control the overall dialogue flow, 
consisting of 146 lines and 4905 characters.
Parts of it are presented in Fig. \ref{f:dfcp} and \ref{f:dfcp-c}.
The part of DFCP shown in Fig. \ref{f:dfcp} defines the dialogue content and flow, 
with the system's utterances and the dialogue flow changing based on the its history.
In this system, GPT-4 generates utterances solely based on the DFCP and dialogue history.
Fig. \ref{f:dfcp-c} shows the part of the DFCP that controls the commands to be executed.
It provides a list of command numbers and their corresponding commands.
The DFCP instructs GPT-4 by determining from the dialogue flow whether to execute any command. 
If necessary, it produces the corresponding command number as output.
For example, if the dialogue goal is achieved, GPT-4 outputs `0' and transitions to closing. 
Thus, this system integrates all instructions related to dialogue history and flow into a single prompt. 
This integration allows for flexible execution of commands 
before and after the system's utterances.




\lstset{
    basicstyle=\fontsize{7}{8}\selectfont\ttfamily,
    breaklines=true,       
    mathescape=true,        
    frame={tb},                           
    numbersep=1.3mm,                        
    lineskip=0mm,                      
    escapechar=| 
}

\begin{figure}[t]
\centering
\begin{lstlisting}
# Constraints
  Essentially|{\rmfamily ,}| follow the sequence of the tasks listed below and execute them one by one.
  You may|{\rmfamily ,}| however|{\rmfamily ,}| change the order of the tasks or omit parts of the tasks|{\rmfamily ,}| depending on the customer|{\rmfamily '}|s requirements.
  $\cdots$
# Task
  1: First|{\rmfamily ,}| ask about the customer|{\rmfamily '}|s hobbies and what they are into these days by saying|{\rmfamily ,}| |{\rmfamily ``}|I|{\rmfamily '}|d like to get to know you better.|{\rmfamily ''}|
  $\cdots$
  2: After telling the customer how glad you are to to hear about the customer|{\rmfamily '}|s hobby|{\rmfamily ,}| please say|{\rmfamily ,}| |{\rmfamily ``}|Let|{\rmfamily '}|s get back to the main topic.|{\rmfamily ''}|
  3: After saying|{\rmfamily ,}| |{\rmfamily ``}|What kind of experiences and memories do you want to create in Kyoto is the key|{\rmfamily ,}||{\rmfamily ''}| ask them what kind of experiences they would like to have on their trip in Kyoto.
  $\cdots$
  10: Confirm with the customer the travel plans determined in this dialog.
# Requirements for sightseeing spots
  $\cdots$
\end{lstlisting}
\caption{Part of dialogue flow control prompt (DFCP)}

\label{f:dfcp}
\end{figure}

%
%

\lstset{
    basicstyle=\fontsize{7}{8}\selectfont\ttfamily, 
    breaklines=true,       
    mathescape=true,        
    frame={tb},                           
    numbersep=1.3mm,                        
    lineskip=0.2mm,                      
    escapechar=|
}

\begin{figure}[t]
\centering
\begin{lstlisting}
Before you speak|{\rmfamily ,}| always decide whether or not to execute the following |{\rmfamily ``}|Command-List.|{\rmfamily ''}|
If you think it is necessary|{\rmfamily ,}| select a command from the list and output only a single digit number.
The command is automatically executed when a single digit number is output.
# Command-List
  0: All phases of the task have been completed and the conversation has settled down|{\rmfamily ,}| so it is terminated.
  1: The first and second sightseeing spots have been decided|{\rmfamily ,}| a restaurant has been suggested|{\rmfamily ,}| and the conversation has come to a natural stop|{\rmfamily ,}| so we will be finalizing the plan.
  2: A customer has asked you to show the atmosphere and pictures of sightseeing spots|{\rmfamily ,}| so you try to show pictures of the sightseeing spots.
  3: Propose a plan.
# When to execute the |{\rmfamily ``}|Command-List|{\rmfamily ''}|
  $\cdots$
\end{lstlisting}
\caption{Example of command control in DFCP}
\label{f:dfcp-c}
\end{figure}

%
%


\subsection{Turn-Take Control Prompt (TTCP)}

The TTCP is a prompt designed to control turn-taking in dialogues,
consisting of 27 lines and 534 characters. An example of TTCP is shown in Fig. \ref{f:ttcp}.
With this prompt described, GPT-4 outputs the command number that it has determined should be executed.
To maintain natural dialogue continuity, 
it is essential to ensure the accurate judgment and control of turn-taking.
Inappropriate control can lead to unwanted barge-ins or unintended silences, causing breakdowns in dialogue.
Cognitive models regarding the conversational counterpart held by users 
is different when interacting with a system as opposed to humans, 
leading to asymmetry in communication as discussed in \cite{Kawakubo23}. 
Consequently, the speech process from the user's side may vary 
when interacting with a human, potentially causing unintended barge-ins and silences.
To resolve such instabilities in turn-taking,
 it is effective to estimate the human cognitive state regarding turn-taking. 
 In our system, we attempted to have GPT-4 estimate 
 the user's cognitive state related to turn-taking based on user or system utterances.
 The prompt controls the operation of the speech recognition system GSR based on these estimations, aiming to suggest interventions to prevent unintended barge-ins and achieve smooth and natural dialogues.
 This kind of control can also contribute to the user's sense of participation in the dialogue \cite{Kawamoto23}.


%
%

\lstset{
    basicstyle=\fontsize{7}{8}\selectfont\ttfamily, 
    breaklines=true,       
    mathescape=true,        
      frame={tb},                           
        numbersep=1.3mm,                        
    lineskip=0mm,                      
    columns=fixed,
    escapechar=|
}

\begin{figure}[t]
\centering
\begin{lstlisting}
The following conditions must be obeyed.
# Constraints
  The customer has spoken|{\rmfamily ,}| but may be about to continue speaking.
  $\cdots$
  Determine whether there is a possibility that the customer will continue speaking. 
  Select from the |{\rmfamily ``}|List of Customer Speech Types|{\rmfamily ''}| below and output a single digit number.
# Output condition
  Output a single digit number corresponding to one of the options from the |{\rmfamily ``}|List of Customer Speech Types.|{\rmfamily ''}|
# Dialog History
  $\cdots$
# List of Customer Speech Types
  0: The customer seems likely to keep talking|{\rmfamily ,}| so it is better for you not to start talking.
  1: The customer may continue talking|{\rmfamily ,}| so it is better for you not to start talking.
  2: The customer may not continue talking|{\rmfamily ,}| so it is acceptable for you to start talking.
  3: The customer seems unlikely to keep talking|{\rmfamily ,}| so it would be better for you to start talking.
  $\cdots$
\end{lstlisting}
\caption{Example of command control in TTCP}
\label{f:ttcp}
\end{figure}

\section{CONCLUSION}

In this paper, we proposed a method for constructing a dialogue system 
that manipulates various aspects of dialogue control, 
a process we refer to as `meta-control,' utilizing LLMs.
We tested this system on the DRC2023 task and 
confirmed that it operates as anticipated.
In principle, it is feasible to integrate the two types of prompts, DFCP and TTCP, 
enabling the system to function with a singular prompt.
The details of this integration will be explored in our future work.
Subtle differences in Japanese expressions in the prompts can greatly affect the GPT-4 output, but it is difficult to describe such subtle differences in English, so we did not mention them in this paper.


\bibliographystyle{IEEEtran}

\end{document}